\documentclass[sigconf, nonacm]{acmart}
\usepackage{tablefootnote}
\usepackage{afterpage}
\usepackage{tabularx}
\usepackage[dvipsnames]{xcolor}
\usepackage{tikz}
\usetikzlibrary{positioning,fit,arrows.meta,backgrounds,calc}

\AtBeginDocument{%
\colorlet{canonical}{teal}
\colorlet{offcanon}{BrickRed}
\colorlet{sharedgray}{gray!15}
\tikzset{
  toolbox/.style={
    rounded corners=3pt, minimum width=3.4cm, minimum height=0.42cm,
    inner sep=4pt, font=\ttfamily\scriptsize, text width=3.3cm,
    draw, align=left
  },
  canonical/.style={toolbox, draw=teal, fill=teal!10, text=teal!80!black},
  offcanon/.style={toolbox, draw=BrickRed, fill=BrickRed!10,
    text=BrickRed!80!black},
  outcomebox/.style={rounded corners=4pt, minimum width=3.4cm,
    minimum height=0.5cm, font=\small\bfseries, align=center, inner sep=4pt},
}%
}

\AtBeginDocument{%
  }

\setcopyright{none}
\renewcommand\footnotetextcopyrightpermission[1]{} 

\acmConference[Review]{Review}{2026}{USA}

\begin{document}

\title[Capable but Unreliable: Agent Failure via Canonical Path Drift]{Capable but Unreliable: Canonical Path Deviation as a Causal Mechanism of Agent Failure in Long-Horizon Tasks}

\author{Wilson Y. Lee}
\email{wilson.yenhsun.lee@gmail.com}
\affiliation{%
  \institution{Independent Researcher}
  \country{United States}
}

\renewcommand{\shortauthors}{Lee}
\acmArticleType{Research}
\acmContributions{WYL designed the study, analyzed the results, and wrote the manuscript.}
\keywords{LLM agents, tool-use, agent reliability, causal inference, benchmarking, stochastic drift}

\begin{abstract}
Why do language agents fail on tasks they are capable of solving? We argue that many such failures are \emph{reliability failures} caused by stochastic drift from a task's latent solution structure, not capability failures caused by insufficient model knowledge. Every well-defined tool-use task imposes a \emph{canonical solution path} (i.e. a convergent set of tool invocations shared across successful runs) and agent success depends critically on whether a trajectory stays within the operating envelope this path defines. We establish this causally using a natural experiment that holds model capability and task difficulty fixed by construction. We analyze trajectories from the Toolathlon benchmark~\citep{toolathlon2025}: 22 frontier models each attempt 108 real-world tool-use tasks across 3 independent runs, yielding 515 model$\times$task units where the \emph{same model} succeeds on some runs and fails on others due to LLM sampling stochasticity alone. Within these units, successful runs adhere significantly more closely to the canonical solution path than failed runs of the same model on the same task ($+$0.060 Jaccard, $p<0.0001$, $n=488$ units, 95\% CI [+0.043,~+0.077]). This result survives six robustness checks including cross-model-family leave-one-out validation, which rules out both circularity and family-level confounding. Critically, the causal mechanism is \emph{gradual and self-reinforcing}: the adherence gap is statistically indistinguishable from zero through the first 50\% of the trajectory, ruling out early-branching selection bias, and each off-canonical tool call raises the probability that the \emph{next} call is also off-canonical by 22.7 percentage points ($\hat{\beta}=+0.227$, $p<0.0001$), more than doubling the baseline off-canonical rate. These findings imply that agent reliability cannot be improved by capability scaling alone, but offer a highly actionable intervention: a simple monitor that restarts the bottom tercile of runs based on mid-trajectory canonical adherence lifts success rates by $+$8.8 percentage points among intervened runs.
\end{abstract}


\maketitle

\begin{figure*}[tp]
  \centering
  \includegraphics[width=\textwidth]{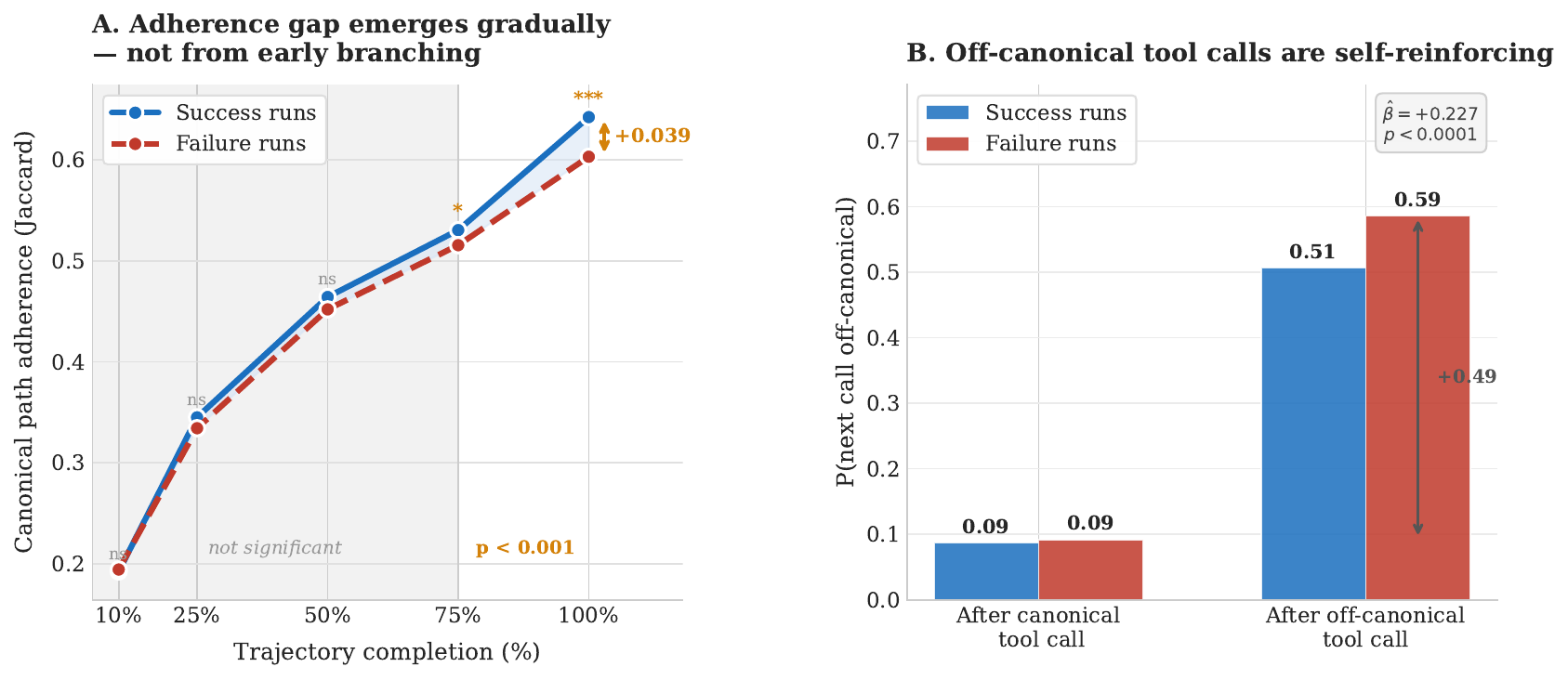}
  \caption{
    \textbf{Canonical path drift is gradual and self-reinforcing.}
    \textbf{(A)} Within mixed-outcome units ($n=488$), the adherence gap
    between successful and failed runs is statistically indistinguishable
    from zero at 10\%, 25\%, and 50\% of trajectory completion, becoming
    significant only at 75\% ($p = 0.043$) and strongly significant at
    full completion ($p < 0.0001$), ruling out early branching as the
    mechanism.
    \textbf{(B)} Conditional transition probabilities show that
    off-canonical calls are self-reinforcing: following an off-canonical
    call, the probability that the \emph{next} call is also off-canonical
    rises by $+22.7$pp relative to following a canonical call
    ($\hat{\beta}=+0.227$, $p<0.0001$), and this compounding is
    significantly stronger in runs that ultimately fail.
    Together, panels A and B establish that agent failure is a
    \emph{gradual compound process}, not a single pivotal mistake.
  }
  \label{fig:drift}
\end{figure*}

\section{Introduction}
\label{sec:intro}

\textbf{Why do language agents fail?}
The dominant answer in the literature is capability: agents fail because they lack the knowledge, reasoning ability, or tool-use skill to solve the task~\citep{yao2023react,shinn2023reflexion,wang2024survey}. This framing has driven the field's response---better models, longer context windows, improved reasoning chains, more sophisticated prompting. Yet a puzzling empirical regularity challenges this account. When the same frontier model attempts the same task multiple times under identical conditions, it sometimes succeeds and sometimes fails. A model that cannot solve a task should fail consistently; a model that can should succeed consistently. The existence of \emph{mixed outcomes}---the same capable model succeeding on one run and failing on another---suggests that something other than capability is determining success on a substantial fraction of agent failures.

We propose that many agent failures are \textbf{reliability failures}, not capability failures. LLM agents are stochastic systems: temperature-based sampling introduces randomness at every decision point, causing the same model to produce different tool sequences across runs. When this stochasticity pushes a trajectory away from the task's solution structure, failure results---not because the model lacks capability, but because this particular draw from its distribution happened to diverge from the region of behavior that leads to success. This distinction matters: capability failures call for better models, but reliability failures call for better constraints on stochastic behavior during execution.

The central concept we introduce is the \textbf{canonical solution path}. For any well-defined task solvable by tool-using agents, successful runs across different models tend to converge on similar tool-use strategies---a set of tools whose invocation is characteristic of task completion. We operationalize this convergent structure as the \emph{empirical consensus tool set}: tools used in more than 50\% of successful runs across models. This set is the operationalization of the canonical solution path---the formal construct we use to measure how closely a trajectory adheres to the task's dominant successful strategy. This convergence is a property of the task, not the model. It emerges from the information dependencies imposed by the task environment: files must be read before they can be edited; search results must be retrieved before they can be synthesized; data must be validated before it can be submitted. These natural orderings of information dependencies create a limited set of viable tool sequences---the canonical path. We show empirically that 58\% of tasks in a large-scale agent benchmark exhibit strong canonical paths (mean Jaccard similarity $>$0.6 among successful runs across different models).

We provide within-unit causal evidence that canonical path adherence causally predicts success within mixed-outcome units, under a standard exogeneity assumption. Our identification strategy exploits a natural experiment (an unplanned source of variation that approximates random assignment) embedded in the Toolathlon benchmark~\citep{toolathlon2025}, which contains 22 frontier models each attempting 108 real-world tool-use tasks across 3 independent runs. Among the 515 model$\times$task units where outcomes vary across runs---pairs where the same model succeeds on some runs and fails on others---variation in outcomes cannot be attributed to model capability or task difficulty, both of which are held fixed within units. The only source of variation is LLM sampling stochasticity. Within these units, successful runs adhere significantly more closely to the canonical solution path than failed runs of the same model on the same task ($+$0.060 Jaccard, $p<0.0001$, $n=488$ units, 95\% CI [+0.043,~+0.077]). This result holds under six robustness checks: leave-one-out canonical definition, cross-family model exclusion, restriction to domain-generic tools only, trajectory-length controls, and leave-one-task-out sensitivity analysis. Under the identifying assumption that run-level stochasticity is exogenous to potential outcomes---plausible given that runs are independent draws from the same model distribution on identical inputs---this within-unit difference is causally identified.

Two additional findings characterize the mechanism. First, canonical path divergence is not a discrete early-branching event but a gradual cumulative process: the adherence gap between successful and failed runs is statistically indistinguishable from zero at 10\%, 25\%, and 50\% of the trajectory, becoming significant only after 75\% completion. This rules out accounts in which a single pivotal wrong decision early in the trajectory determines failure. Instead, agents gradually drift from the canonical path over the course of execution, with each off-canonical tool call subtly shifting the information state in ways that make subsequent canonical selections less likely. We use ``mechanism'' in the causal sense: a causally-identified pathway from stochastic deviation to failure, characterised by its gradual dynamics and self-reinforcing structure; a formal information-theoretic account of \emph{why} off-canonical calls compound is left to future work. Second, the effect is broadly consistent across model families: 10 of 11 families tested show positive within-unit adherence gaps, with 5 reaching individual significance despite modest within-family sample sizes.

Our findings carry three implications. For agent \emph{deployment}, reliability cannot be improved by capability alone---constraining stochastic drift toward canonical paths throughout execution is a complementary and underexplored intervention. We estimate that a monitor flagging the bottom tercile of runs by canonical path adherence at 75\% trajectory completion and restarting them lifts success rates by $+8.8$pp among intervened runs (Section~\ref{sec:discussion}), without any change to the underlying model. For \emph{benchmark design}, single-run evaluations conflate capability and reliability, potentially misattributing reliability failures to capability gaps; benchmarks should report variance across runs alongside mean performance. For our \emph{theoretical understanding} of agent failure, the gradual drift mechanism suggests that mid-trajectory monitoring and correction---detecting canonical path deviation before it compounds---may be more effective than initialization-time interventions such as better prompting or chain-of-thought.

\section{Related Work}
\label{sec:related}

\paragraph{LLM agent benchmarks.}
A growing body of work evaluates LLM agents on real-world tool-use tasks~\citep{yao2023webagent,zhou2023webarena,yang2024swebench,
liu2023agentbench}. These benchmarks primarily measure \emph{capability}---typically reported as pass@1 or pass@$k$---and treat failures as evidence of insufficient model ability. A notable exception is $\tau$-bench~\citep{yao2024taubench}, which introduces the P\^{}$k$ metric (the probability of succeeding on all $k$ independent trials) to surface the reliability gap between what models can occasionally achieve and what they achieve consistently. $\tau$-bench establishes that the reliability gap is real and large; our work provides the causal mechanism behind it, showing that stochastic deviation from canonical solution paths is the structural source of the within-unit inconsistency that P\^{}$k$ measures. Our work reframes a substantial fraction of benchmark failures as reliability failures: the Toolathlon benchmark~\citep{toolathlon2025} reveals that 22.5\% of model$\times$task pairs exhibit mixed outcomes across runs, indicating that models are capable of solving these tasks but fail to do so consistently.

\paragraph{LLM stochasticity and sampling.}
The stochastic nature of LLM sampling has been studied primarily as a resource to exploit rather than a failure mechanism to mitigate. Work on self-consistency~\citep{wang2023selfconsistency} and majority-vote decoding uses sampling diversity to improve accuracy by aggregating across runs. Pass@$k$ metrics~\citep{chen2021codex} similarly treat multiple runs as a capability probe. In contrast, we treat stochasticity as a source of reliability failures and study its structural consequences for trajectory quality. To our knowledge, no prior work characterizes the mechanism by which stochastic sampling causes capable agents to fail.

\paragraph{Defining correct execution.}
A complementary line of work defines correct agent execution through expert-designed reference plans. AssetOpsBench~\citep{patel2025assetopsbench}, a large-scale industrial benchmark spanning 600+ tools across Industry~4.0 environments, evaluates agents against human-authored expert scenarios that specify the correct reasoning and action sequence for each task. Our approach differs in deriving canonical paths empirically from the consensus of successful runs across models via CF-LOO, rather than requiring expert annotation, making the framework scalable to new tasks without domain expertise.

\paragraph{Agent failure analysis.}
Several papers have analyzed failure modes in LLM agents~\citep{liu2023lost,yang2024failures,ruan2023identifying}. This work is primarily taxonomic---classifying failures into categories such as reasoning errors, tool misuse, and context loss---and largely descriptive. Our contribution is causal: we identify a specific mechanism (canonical path deviation) and provide causally-identified evidence that it determines success within model$\times$task units.

\paragraph{Planning and task structure.}
The connection between LLM agents and classical AI planning has been noted in several works~\citep{valmeekam2023planning, kambhampati2024llms,silver2024generalized}. \citet{newell1972human}'s problem space hypothesis---that all problem-solving can be characterized as search through a state space---provides the theoretical grounding for our canonical path concept. We extend this tradition by providing the first empirical operationalization of solution paths for LLM tool-use tasks and demonstrating their causal relevance.

\paragraph{Reliability engineering.}
The distinction between capability and reliability is well-established in engineering~\citep{perrow1984normal,hollnagel2004barriers}. \citet{perrow1984normal}'s Normal Accident Theory identifies how stochastic perturbations in complex systems compound into failures even when individual components are capable. We apply this lens to LLM agents, showing that stochastic tool selection errors compound gradually over agent trajectories in a manner consistent with Perrow's account of drift-induced failure.

\section{Dataset and Natural Experiment}
\label{sec:data}

\subsection{The Toolathlon Benchmark}

We analyze trajectories from the publicly available Toolathlon benchmark~\citep{toolathlon2025}, which contains trajectories from 22 frontier language models each attempting 108 real-world tool-use tasks across 3 independent runs, yielding approximately 7,000 trajectories. Tasks span diverse domains including web navigation, file manipulation, code execution, data analysis, and multi-step research workflows. Each task provides the same instructions, tool environment, and evaluation criteria across all models and runs. Tools are invoked via structured function calls; each trajectory records the full sequence of tool calls, their arguments, and their outputs. Success is evaluated by automated task-specific criteria.

Toolathlon is distinguished from prior agent benchmarks by three design properties that make it particularly suited to trajectory analysis. First, \textbf{breadth}: the benchmark spans 32 software applications and 604 tools, ranging from everyday platforms (Google Calendar, Notion, Gmail) to professional environments (WooCommerce, Kubernetes, Weights \& Biases, BigQuery), ensuring that findings are not artifacts of a narrow tool ecosystem. Second, \textbf{environmental realism}: rather than synthetic task states, each task is initialized from realistic environment conditions (e.g., Canvas courses with real student rosters, financial spreadsheets with genuine market data) so agents face the same context diversity as deployment settings. Third, \textbf{long-horizon complexity}: tasks require approximately 20 tool-calling turns on average to complete, with many tasks requiring coordinated use of tools across multiple applications in a single trajectory. This combination of breadth, realism, and depth creates the trajectory-length and tool-diversity conditions necessary to observe the gradual behavioral patterns we study. Even the best-performing model (Claude~4.5~Sonnet) achieves only 38.6\% success, leaving substantial outcome variation for within-unit analysis~\citep{toolathlon2025}.

\begin{table}[h]
\centering
\caption{Dataset structure and mixed-outcome sample.}
\label{tab:data}
\small
\begin{tabular}{lr}
\toprule
\textbf{Statistic} & \textbf{Value} \\
\midrule
Total models & 22 \\
Total tasks & 108 \\
Runs per model$\times$task & 3 \\
Total trajectories & 7,003 \\
\midrule
Model$\times$task units (3 runs) & 2,286 \\
Always-succeed units & 286 (12.5\%) \\
Always-fail units & 1,485 (65.0\%) \\
Mixed-outcome units & 515 (22.5\%) \\
\quad Of which: 1/3 success & 267 \\
\quad Of which: 2/3 success & 248 \\
\midrule
Mixed units in canonical analysis & 488\tablefootnote{Of the 515 mixed-outcome units, 488 have valid CF-LOO canonical paths (at least 3 successful runs from model families other than the target unit's family). The remaining 27 units are excluded because their task has no successful runs across all models (11 units), or because successful runs are concentrated within the target unit's model family (16 units), leaving insufficient cross-family evidence to define the CF-LOO canonical path. Results are substantively identical under the less conservative LOO specification ($n=495$; Table~\ref{tab:main}).} \\
\bottomrule
\end{tabular}
\end{table}

\subsection{The Natural Experiment}

The three-run structure of Toolathlon creates a natural experiment (an unplanned source of variation that mimics random assignment) for causal identification. Within each model$\times$task \emph{unit} (a specific model attempting a specific task), runs are independent draws from the same model distribution on identical inputs. The only source of within-unit variation is LLM sampling stochasticity.

We identify 515 \textbf{mixed-outcome units}: model$\times$task pairs where the same model succeeds on at least one run and fails on at least one other run ($22.5\%$ of all units with exactly 3 runs). Within these units, outcome variation cannot be attributed to model capability (held fixed: same model) or task difficulty (held fixed: same task). This within-unit design absorbs all between-unit confounds and isolates the causal effect of run-level behavioral variation on success~\citep{cunningham2021mixtape}. Concretely, this is a matched-pairs comparison: rather than comparing model~A on task~X to model~B on task~Y---where both capability and task difficulty confound any difference---we compare model~A's successful runs on task~X to its own failed runs on the exact same task. Capability and task difficulty cancel out by construction; the only thing that can differ between a success and a failure in the same unit is how the model behaved on that particular run.

Table~\ref{tab:data} summarizes the dataset structure and the
mixed-outcome sample.

\subsection{Identifying Assumption}

Our within-unit estimates are causally identified under the assumption that run-level stochasticity is exogenous to potential outcomes---that is, the random seed governing LLM sampling for run $r$ is independent of the potential success outcomes $(Y_{i,j,1}, Y_{i,j,2}, Y_{i,j,3})$ for model $i$ on task $j$. In plain terms: the randomness that determines which tool sequence the model generates on a given run is not systematically related to whether that sequence will succeed---the model's sampler does not ``know'' in advance which draw will lead to success. This assumption is plausible: runs are generated independently with no shared state, and LLM sampling randomness is determined by hardware random number generators prior to any task-specific computation. We provide three forms of empirical support. First, the distribution of first tool calls is statistically indistinguishable between runs that eventually succeed and runs that eventually fail within mixed units (minority first-tool-choice success rate: 51.3\%, 95\% CI [45.6\%, 57.0\%], $p = 0.69$, $n = 308$ minority-tool runs), confirming that initial sampling is not predictive of outcome. Second, the gradual drift finding (Section~\ref{sec:results}) shows that the adherence gap is statistically indistinguishable from zero at 10\%, 25\%, and 50\% of the trajectory; if stochasticity were correlated with potential outcomes, systematic early differences would be expected but are not observed. Third, the cross-family LOO specification ensures that canonical paths are constructed from entirely independent model families, so any within-family correlation between runs cannot inflate the estimated effect. Our causal claims are local to mixed-outcome units and do not imply that canonical path deviation explains all agent failures.

\section{Canonical Solution Paths}
\label{sec:canonical}

\subsection{Definition}

For a task $j$, let $\mathcal{S}_j$ denote the set of all successful trajectories across all models and runs. For each trajectory $\tau \in \mathcal{S}_j$, let $\mathcal{T}(\tau)$ denote the set of distinct tool names invoked. We define the \textbf{empirical consensus tool set} $C_j$---which we refer to as the canonical tool set for brevity---as:

Three design choices warrant brief justification. We use \emph{unordered sets} rather than sequences because we aim to capture the structural envelope of successful behavior---which tools a trajectory must visit---rather than a specific policy for visiting them; ordered metrics impose sequence constraints that add noise without increasing sensitivity (we verify this in §\ref{sec:results}). We use a \emph{majority consensus threshold} ($>$50\%) because it is the weakest assumption under which a tool can be called convergent: any lower threshold would include tools used by a minority of successful runs, conflating canonical tools with incidental ones. We use \emph{Jaccard similarity} to measure adherence because it penalizes both missing canonical tools and adding non-canonical ones symmetrically, reflecting the dual failure modes of omission and distraction.
\begin{equation}
  C_j = \left\{ t \;\middle|\; \frac{\sum_{\tau \in \mathcal{S}_j}
  \mathbf{1}[t \in \mathcal{T}(\tau)]}{|\mathcal{S}_j|} > 0.5
  \right\}
  \label{eq:canonical}
\end{equation}
That is, the canonical tool set consists of tools used in more than 50\% of successful runs on task $j$. This is an empirical construct---a descriptive summary of convergent successful behavior---and does not claim that these tools are individually necessary or jointly sufficient for task completion. The \textbf{canonical path adherence} of a trajectory $\tau$ on task
$j$ is:
\begin{equation}
  A(\tau, j) = \frac{|\mathcal{T}(\tau) \cap C_j|}{|\mathcal{T}(\tau)
  \cup C_j|}
  \label{eq:adherence}
\end{equation}
This is the Jaccard similarity between the trajectory's tool set and the canonical tool set.

\subsection{Empirical Properties}

We compute canonical tool sets for 74 tasks with sufficient successful runs ($|\mathcal{S}_j| \geq 3$). Figure~\ref{fig:main} panel A shows the distribution of within-unit adherence effects across 70 tasks; 56 of 70 (80\%) show positive effects.

Tasks vary substantially in canonical path strength---the mean pairwise Jaccard similarity among successful runs on that task. Fifty-eight percent of tasks exhibit strong canonical paths (mean Jaccard $> 0.6$), indicating high convergence among successful strategies. Canonical path strength is predicted by canonical path size ($r = +0.37$, $p < 0.01$) but not by overall task success rate ($r = -0.01$, $p = 0.92$), confirming that canonical paths reflect task structure rather than task difficulty. Canonical paths are predominantly generic: only 13.5\% of canonical tools on average contain task-domain identifiers, indicating that canonical paths capture general solution strategies, not trivially obvious tool choices.

\subsection{Cross-Family Validation}

To ensure that canonical paths are genuine task-level properties and not artifacts of the behavior of any particular model family (i.e., same organization or base model lineage), we employ a
\textbf{cross-family leave-one-out} (CF-LOO) definition. For a trajectory from model family $f$ on task $j$, the CF-LOO canonical set excludes all successful runs from family $f$:
\begin{equation}
  C_j^{-f} = \left\{ t \;\middle|\; \frac{\sum_{\tau \in
  \mathcal{S}_j^{-f}} \mathbf{1}[t \in \mathcal{T}(\tau)]}
  {|\mathcal{S}_j^{-f}|} > 0.5 \right\}
  \label{eq:cfloocanonical}
\end{equation}
where $\mathcal{S}_j^{-f}$ excludes all successful runs from family $f$. This definition is used as the primary specification in all causal
analyses below.

\section{Results}
\label{sec:results}

\subsection{Main Result: Canonical Path Adherence Predicts Success}

\paragraph{Specification.}
For each mixed-outcome unit $(i, j)$, we compute the mean CF-LOO canonical path adherence of successful runs $\bar{A}^+_{ij}$ and failed runs $\bar{A}^-_{ij}$. The within-unit adherence gap is $\Delta_{ij} = \bar{A}^+_{ij} - \bar{A}^-_{ij}$. We test $H_0: \mathbb{E}[\Delta_{ij}] = 0$ using a paired $t$-test and bootstrap confidence intervals (5,000 resamples).

\paragraph{Result.}
Successful runs adhere significantly more closely to the canonical solution path than failed runs of the same model on the same task. The mean within-unit adherence gap is $+0.060$ Jaccard ($p < 0.0001$, $n = 488$ units, 95\% CI [+0.043,~+0.077]). In 59.8\% of mixed-outcome units, the successful run is more adherent than the average failed run.

\paragraph{Effect size interpretation.}
To translate the $+0.060$ Jaccard gap into practical terms, we regress success on within-unit demeaned canonical adherence using logistic regression across all mixed-outcome trajectories ($n = 1{,}476$). A $+0.060$ increase in within-unit adherence implies a $+5.3$pp increase in success probability (odds ratio $= 1.24$, Cohen's $d = 0.48$, medium effect). Given that baseline success rates in mixed units range from 33\% to 67\% by definition, a $+5.3$pp shift represents a meaningful reliability improvement---achievable without any change to the underlying model.

\begin{figure*}[t]
  \centering
  \includegraphics[width=\textwidth]{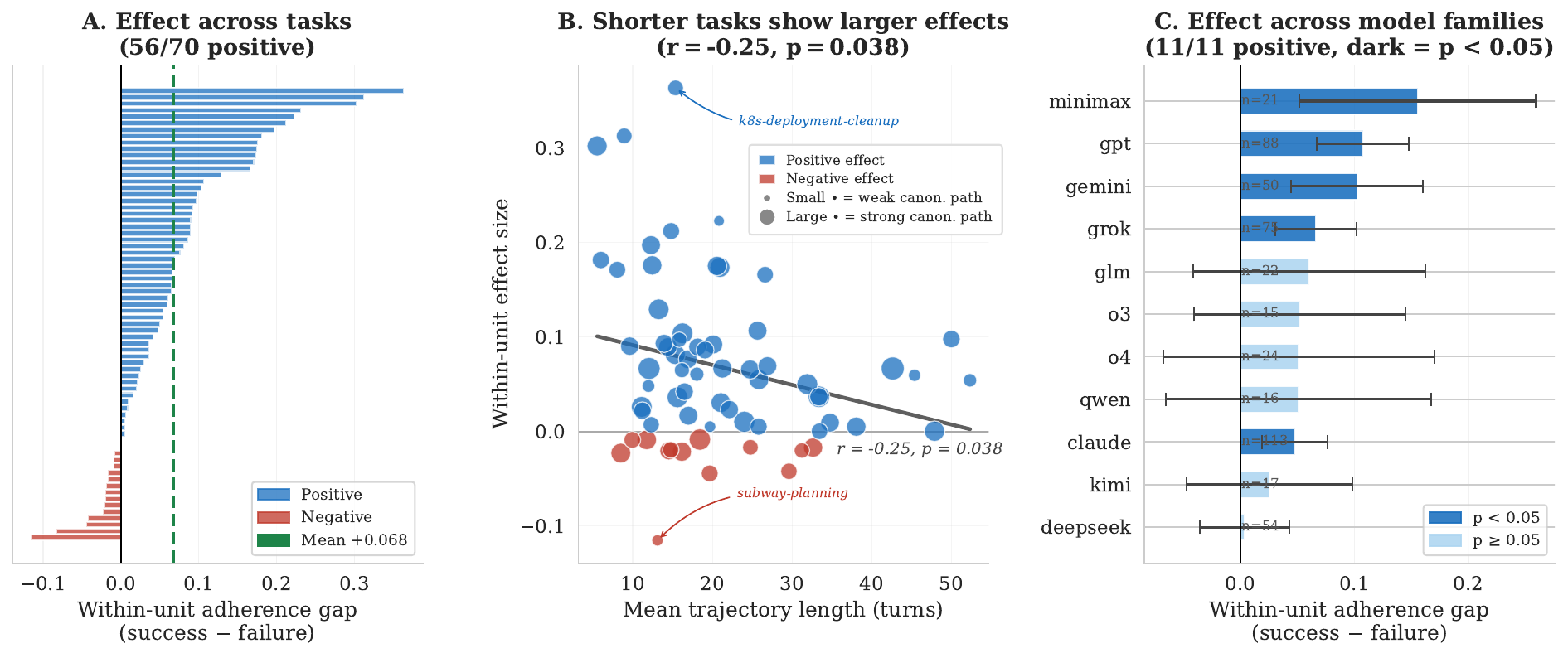}
  \caption{
    \textbf{Robustness and generalization of the canonical path
    adherence effect.}
    \textbf{(A)} Within-unit adherence gaps across 70 tasks; 56/70
    (80\%) show positive effects (mean $= +0.068$, green dashed line).
    \textbf{(B)} Effect size is larger for shorter tasks
    ($r = -0.25$, $p = 0.038$), consistent with less opportunity for
    course-correction after deviation; colour indicates canonical path
    strength.
    \textbf{(C)} All 11 model families show positive gaps; 5 reach
    individual significance ($p < 0.05$, dark blue).
    Error bars show $\pm 1.96$ SE.
  }
  \label{fig:main}
\end{figure*}

Table~\ref{tab:main} presents the main result alongside all robustness specifications. Figure~\ref{fig:main} panel C shows the result broken down by model family.

\begin{table}[t]
\centering
\caption{Main result and robustness checks.
The \emph{within-unit adherence gap} is the difference in canonical path adherence (Jaccard similarity) between successful and failed runs of the \emph{same model} on the \emph{same task}, averaged across all mixed-outcome units (model$\times$task pairs where outcomes vary across runs). A positive gap indicates that successful runs adhere more closely to the canonical tool set than failed runs within the same unit, isolating the causal effect of trajectory-level variation on success while holding model capability and task difficulty fixed. All specifications use paired within-unit comparisons; CF-LOO = cross-family leave-one-out canonical path definition.
$^{***}p<0.001$.}
\label{tab:main}
\small
\begin{tabular}{lcccc}
\toprule
\textbf{Specification} & \textbf{Gap} & \textbf{CI} & \textbf{$N$} &
\textbf{Sig.} \\
\midrule
Original (naive) & +0.082 & [+0.059,~+0.105] & 505 & $^{***}$ \\
Leave-one-out & +0.067 & [+0.050,~+0.084] & 495 & $^{***}$ \\
CF-LOO (headline) & +0.060 & [+0.043,~+0.077] & 488 & $^{***}$ \\
Generic tools only & +0.036 & [+0.022,~+0.051] & 473 & $^{***}$ \\
Length-residualized & +0.053 & [+0.037,~+0.071] & 481 & $^{***}$ \\
Leave-one-task-out & +0.050 & [+0.033,~+0.066] & 471 & $^{***}$ \\
\bottomrule
\end{tabular}
\end{table}

\subsection{Robustness Checks}

We test six alternative explanations for the main result.

\paragraph{R1: Circularity.}
If a model's successful run contributes to its own canonical path definition, adherence is trivially inflated for that run. We address this with the leave-one-out (LOO) specification, which defines the canonical path excluding the target model's runs entirely. The LOO estimate (+0.067) closely matches the naive estimate (+0.082), indicating that circularity explains only $\approx$18\% of the naive effect.

\paragraph{R2: Family bias.}
If model families share behavioral tendencies, canonical paths derived from one family may artificially inflate adherence for models from that family. The CF-LOO specification addresses this by excluding the entire model family from the canonical path definition. The CF-LOO estimate (+0.060) survives this stricter exclusion, and the effect is positive for all 11 model families tested (Figure~\ref{fig:main}C).

\paragraph{R3: Domain-obvious tools.}
Canonical paths might trivially encode domain-specific tools (e.g., \texttt{notion-*} tools for Notion tasks) that any non-confused model would use. We test this by restricting canonical paths to tools whose names do not contain task-domain identifiers. The generic-tools estimate (+0.036, 95\% CI [+0.022,~+0.051]) remains significant at $p < 0.0001$, confirming that the effect is not driven by trivially obvious domain tool choices.

\paragraph{R4: Trajectory length confound.}
Longer trajectories mechanically include more tools, potentially inflating Jaccard similarity with any reference set. We address this by residualizing adherence on trajectory length (total turns) within units before testing. The length-residualized estimate (+0.053, 95\% CI [+0.037,~+0.071]) confirms that the effect is not a proxy for trajectory length.

\paragraph{R5: Task leverage.}
The aggregate result could be driven by a small number of high-effect tasks. We test this by iteratively removing the most influential task and re-estimating. Removing the most influential task (\texttt{set-conf-cr-ddl}) changes the estimate minimally (+0.050, 95\% CI [+0.033,~+0.066]), and 56 of 70 tasks show positive effects.

\paragraph{R6: Canonical threshold sensitivity.}
The canonical tool set uses a majority threshold ($>$50\% of successful runs; Eq.~\ref{eq:canonical}). We re-estimate the within-unit adherence gap at thresholds of 40\%, 50\%, 60\%, and 70\%. Effects range from $+0.055$ to $+0.062$ Jaccard across all four thresholds, all significant at $p < 0.0001$, with the fraction of positive units ranging from 55.7\% to 59.4\%. The finding is insensitive to reasonable threshold choices.

\subsection{Falsification of Early Branching; Evidence for Gradual Drift}

A natural alternative account of the main result is the \emph{early-branching hypothesis}: a single wrong tool call early in the trajectory deterministically branches the run away from the canonical path, causing failure. This hypothesis makes a clear prediction---the adherence gap between successful and failed runs should be large and detectable from the first tool calls onward. We test this prediction directly.

We measure the within-unit adherence gap at five trajectory completion points: 10\%, 25\%, 50\%, 75\%, and 100\%. Figure~\ref{fig:drift} shows the results. The early-branching prediction is not supported: the adherence gap is statistically indistinguishable from zero at 10\% ($\Delta = -0.002$, $p = 0.74$), 25\% ($\Delta = +0.011$, $p = 0.12$), and 50\% ($\Delta = +0.012$, $p = 0.09$) of the trajectory. The gap becomes marginally significant only at 75\% ($\Delta = +0.015$, $p = 0.043$) and strongly significant at full completion ($\Delta = +0.039$, $p < 0.0001$). The monotonically increasing gap is instead consistent with \emph{gradual cumulative drift}: each off-canonical tool call slightly shifts the information state, making subsequent canonical selections less likely and compounding over the trajectory.

We further stress-test the early-branching hypothesis using a difference-in-differences (DiD) design that treats the first off-canonical tool call as a discrete treatment event. If early branching were the mechanism, runs that deviate earlier should fail at higher rates. The DiD design fails the parallel trends assumption---runs that deviate early already show lower pre-deviation adherence than runs that deviate late ($\Delta = -0.158$, $p < 0.0001$)---indicating that early deviators were already drifting continuously before their first off-canonical call, not branching discretely. The DiD estimate is not significant ($\Delta = +0.039$, $p = 0.36$), and there is no dose-response relationship between first-deviation timing and success ($r = -0.020$, $p = 0.61$). All three tests falsify the early-branching hypothesis and support gradual drift as the operative mechanism.

\paragraph{Self-reinforcing drift.}
The gradual drift account implies a specific micro-level mechanism: each off-canonical tool call should increase the probability that the \emph{next} call is also off-canonical, as each deviation shifts the information state in ways that make canonical selections less likely. We test this directly using a repeated-treatment specification. For each consecutive tool-call pair $(t, t+1)$ within a trajectory, we regress an indicator for the call at $t+1$ being off-canonical on an indicator for the call at $t$ being off-canonical, with trajectory fixed effects to absorb baseline drift tendencies.

The estimated self-reinforcing coefficient is $\hat{\beta} = +0.227$ ($p < 0.0001$, clustered 95\% CI [+0.202,~+0.252], $n = 35{,}045$ consecutive tool-call pairs from mixed-outcome units). An off-canonical call at position $t$ increases the probability that position $t+1$ is also off-canonical by 22.7pp, more than doubling the baseline rate of 18.1\%. The effect is significantly stronger in runs that eventually fail ($\hat{\beta} = +0.247$) than in runs that eventually succeed ($\hat{\beta} = +0.207$), consistent with self-reinforcing drift compounding all the way to failure in the former and being partially arrested through recovery in the latter. This provides direct micro-level evidence for the compounding mechanism underlying the aggregate adherence gap.

\subsection{Cross-Family Universality}

Figure~\ref{fig:main} panel C shows within-unit adherence gaps for all 11 model families. All 11 families show non-negative gaps, and 5 reach individual significance (claude $p = 0.0015$; gpt $p < 0.0001$; gemini $p = 0.0011$; grok $p = 0.0006$; minimax $p = 0.0082$). Families that do not reach individual significance (deepseek, kimi, o3, o4, qwen, glm) are predominantly explained by small within-family sample sizes ($n = 15$--$54$); for five of the six, this is the likely explanation. The exception is deepseek ($n = 54$, the largest non-significant family), which shows a near-zero point estimate ($+0.004$), suggesting either genuine attenuation or idiosyncratic behavior in this family. The remaining five non-significant families show positive estimates ranging from $+0.025$ (kimi) to $+0.060$ (glm), consistent in magnitude with the five significant families, which range from $+0.048$ (claude) to $+0.155$ (minimax). The near-universal pattern of positive effects across 10 of 11 families, spanning diverse training paradigms and model architectures, supports the interpretation that canonical path adherence reflects a task-level structural property rather than a model-family artifact.

\subsection{Variance Signature of Reliability vs.\ Capability Failure}

The within-unit adherence variance across unit types provides independent support for the reliability failure account. If mixed-outcome units represent reliability failures caused by stochastic drift, they should show higher within-unit adherence variance than always-succeed units, where the model consistently finds the canonical path, and potentially higher than always-fail units, where the model consistently takes the wrong path.

We compute the within-unit standard deviation of CF-LOO canonical path adherence across all three runs for each unit type. Always-succeed units show significantly lower adherence variance than mixed units (std $= 0.086$ vs.\ $0.109$, $\Delta = -0.023$, $p = 0.0001$, Cohen's $d = -0.30$): consistent success is associated with consistent path-following. Always-fail units show intermediate variance (std $= 0.099$), lower than mixed units ($p = 0.038$) but higher than always-succeed units ($p = 0.007$). This ordering---always-succeed $<$ always-fail $<$ mixed---is consistent with a two-failure-type interpretation: always-fail units represent models that consistently take the wrong path (low variance in failure, a capability signature), while mixed units are genuinely at the stochastic boundary where sampling variation determines outcome (high variance, a reliability signature). The proportion of ``locked-in'' units where all three runs follow nearly identical tool sequences (std $< 0.05$) further supports this pattern: 33.0\% of always-succeed units, 25.4\% of always-fail units, and 23.1\% of mixed units.

\section{Task-Level Analysis}
\label{sec:task}

\subsection{What Predicts Canonical Path Strength?}

We examine what structural properties of tasks predict the strength of their canonical paths (mean Jaccard similarity among successful runs). Canonical path strength is significantly predicted by canonical path size ($r = +0.37$, $p < 0.01$): tasks with larger canonical tool sets show greater convergence among successful runs. Mean trajectory length ($r = +0.00$, $p = 0.99$) and overall success rate ($r = -0.01$, $p = 0.92$) do not predict canonical path strength, confirming that canonical paths reflect task solution structure independent of task difficulty.

\subsection{What Predicts Effect Size?}

Panel B of Figure~\ref{fig:main} shows that the within-unit adherence effect is larger for shorter tasks ($r = -0.25$, $p = 0.038$) and tasks with fewer unique tools ($r = -0.30$, $p = 0.014$). This is consistent with the gradual drift mechanism: in shorter tasks with simpler tool environments, there is less opportunity for course-correction after canonical path deviation, so early drift has larger consequences. Conversely, longer tasks with richer tool environments offer more recovery opportunities, attenuating the adherence effect. We note that agents retry the same failed tool in approximately 47\% of error events regardless of whether the run eventually succeeds, suggesting that retry behavior is a fixed response pattern rather than a reliability-relevant decision point.

\section{Discussion}
\label{sec:discussion}

\subsection{Theoretical Implications}

Our results support a \textbf{reliability engineering} account of LLM agent failure. Classical reliability theory~\citep{perrow1984normal} distinguishes between failures caused by insufficient capability and failures caused by stochastic perturbations that push otherwise-capable systems outside their operating envelope. We show that LLM agent failures exhibit the latter pattern: the same capable model produces mixed outcomes across runs because stochastic sampling occasionally produces trajectories that drift outside the canonical solution envelope.

The gradual drift mechanism connects to \citet{newell1972human}'s problem space hypothesis: agents search a task-defined problem space, and canonical paths define the productive region of that space. Stochastic sampling introduces a random walk component into this search; when the walk drifts too far from the canonical region, the probability of recovery decreases monotonically, producing the compounding divergence we observe empirically. The repeated-treatment estimate ($\hat{\beta} = +0.227$) gives this random walk a precise drift coefficient: each step off the canonical path increases the probability of the next step also being off-canonical by 22.7pp, quantifying the rate at which the information state degrades following a deviation.

The variance analysis (Section~\ref{sec:results}) further grounds this account in an empirical taxonomy of failure types. Always-succeed units show low adherence variance: the model reliably finds the canonical path, a capability-plus-reliability success. Always-fail units show intermediate variance: the model consistently takes the wrong path, implicating a capability gap that path guidance alone cannot fix. Mixed units show the highest variance: the model is capable but stochastically unreliable, the target population for reliability interventions. This taxonomy is directly observable from repeated-run benchmark data and provides a principled basis for triaging agent failures into those that require better models versus those that require better execution constraints.

\subsection{Practical Implications}

\paragraph{Canonical path scaffolding.}
Our results suggest that reliability can be improved by guiding agents toward canonical paths during execution. Because canonical paths can be extracted from successful runs without requiring task-specific expert knowledge, this approach scales to new tasks automatically once a sufficient number of successful runs have been collected---no manual annotation or domain expertise is needed beyond running the agent. In Toolathlon, canonical paths stabilize with 3–5 successful runs from diverse models; tasks with fewer successes show higher cross-run variance in tool sets but still converge on a recognizable core. We provide a counterfactual estimate of the potential lift. For each trajectory in our mixed-outcome sample, we compute the predicted success probability at observed adherence and at the adherence level achieved by successful runs of the same model on the same task (the conservative intervention target). A monitor that flags the bottom tercile of runs by canonical path adherence at 75\% of trajectory completion---the earliest point at which drift is statistically detectable---and restarts them is expected to lift success rates by $+8.8$pp among intervened runs (95\% CI [$+7.6$pp, $+9.9$pp]), improving their success rate from 44.9\% to 53.7\% without any change to the underlying model. More aggressive flagging thresholds yield larger per-run lifts at the cost of intervening on a smaller fraction of runs: flagging only runs that deviate more than 10\% below their unit mean yields an expected $+13.1$pp lift (95\% CI [$+10.9$pp, $+15.3$pp]) on 11.7\% of runs. These estimates assume that a restarted run achieves the adherence level of successful runs on the same task---a conservative target that does not require perfect adherence, only recovery to the level empirically observed in successful runs.

\paragraph{Mid-trajectory intervention.}
The gradual drift finding implies that interventions are most valuable mid-trajectory, not at initialization. Prompting interventions (chain-of-thought, task decomposition) operate at initialization and may be insufficient for the reliability problem we identify. Runtime scaffolding that monitors adherence and intervenes when drift is detected is a more targeted approach.

\paragraph{Reliability-aware benchmarking.}
Aggregate pass rate conflates capability (can the model solve this task?) and reliability (does it solve it consistently?). Two models with identical aggregate pass rates can have entirely different failure profiles: one that succeeds 3/3 on one-third of tasks and fails the rest has a clean capability boundary, while one that succeeds 1/3 on every task fails unpredictably on tasks it is demonstrably capable of solving. The P\^{}$k$ metric~\citep{yao2024taubench} addresses this by measuring the probability of succeeding on all $k$ trials, penalizing inconsistency directly and providing an aggregate reliability floor. We propose that benchmarks also report the fraction of \emph{mixed-outcome units}, which answers a distinct and complementary question: not how reliable a model is overall, but how much of its unreliability is \emph{addressable}. P\^{}$k$ cannot distinguish between always-fail units (capability gaps, where no reliability intervention will help) and mixed-outcome units (drift-induced failures, where canonical path scaffolding can recover success). The mixed-outcome fraction isolates the latter population---the upper bound on what mid-trajectory interventions can recover---and is therefore the natural denominator for evaluating the impact of reliability interventions. In Toolathlon, 22.5\% of model$\times$task units are mixed-outcome, representing the precise scope of the $+8.8$pp lift estimated in this section: not all failures, only the recoverable ones. We apply this metric to all 22 Toolathlon models in Appendix~\ref{app:reliability_metrics}, where MO/P\^{}$k$ reveals substantial variation in intervention scope that P\^{}$k$ alone cannot surface.

\paragraph{Deployment implications of unreliability.}
Reliability failures are more costly in production than capability failures: a model that fails unpredictably on tasks it can solve generates retries, consuming latency and compute beyond what its benchmark score implies, with no principled basis for knowing when to stop retrying. Capability failures, by contrast, are predictable and can be triaged. The 22.5\% of model$\times$task units that are mixed-outcome are the highest-value deployment target: the model is capable, the canonical path exists, and mid-trajectory monitoring can recover success without any change to the underlying model.

\paragraph{Benchmarks as natural experiments.}
Large-scale agent benchmarks are typically analyzed as capability scorecards, with pass rates aggregated across models and tasks to rank performance. We argue they also contain latent natural experiments that are largely untapped. The identification strategy here---exploiting outcome variance within model$\times$task units where the same model succeeds and fails across runs---is only possible because the Toolathlon benchmark was designed with multiple runs per unit. This structure, common in several existing benchmarks, embeds a quasi-experimental design that holds model capability and task difficulty fixed and isolates run-level variation as the sole source of outcome differences. Such designs are not a fallback for when purpose-built experiments are unavailable; they offer identification guarantees that a purpose-built single-run experiment could not replicate, because the within-unit counterfactual is only credible when the same model has been observed to both succeed and fail on the same task. We encourage benchmark designers to treat repeated runs not merely as variance-reduction tools but as the raw material for causal identification.

\subsection{Limitations}

\paragraph{Three runs per unit.}
The Toolathlon benchmark provides exactly 3 runs per model$\times$task unit. This is sufficient for identifying mixed-outcome units and computing within-unit comparisons, but limits the precision of per-unit estimates. Future work should examine whether the canonical path effect strengthens with larger run counts.

\paragraph{Jaccard similarity.}
Our canonical path adherence metric uses unordered tool sets (Jaccard similarity) rather than ordered sequences. Sequence-based metrics (e.g., edit distance, bigram overlap) may capture additional structure. We verified that sequence-based metrics produce qualitatively similar but statistically weaker results, likely due to sparser within-unit variation at the sequence level.

\paragraph{Tool granularity.}
We define canonical paths at the level of tool names (e.g., \texttt{filesystem-read\_file}). Finer-grained analysis of tool arguments could reveal additional solution structure, but increases sparsity substantially.

\paragraph{Multi-modal solution paths.}
Our canonical path definition identifies tools used in $>$50\% of successful runs, which implicitly assumes that successful runs converge on a single dominant strategy. For tasks where multiple equally-valid but disjoint solution paths exist---for example, tasks solvable via either a web-search route or a direct API route---the empirical consensus tool set will be sparse, capturing only tools common to both strategies. This attenuates the Jaccard adherence signal toward zero, making our test conservative for multi-modal tasks rather than inflated: if the canonical path mechanism is detectable despite this attenuation, the true effect in unimodal tasks is likely larger. Future work could extend the framework using mixture models or hierarchical path clustering to explicitly represent multi-modal solution structures.

\paragraph{Toward experimental validation.}
The present paper establishes the canonical path deviation mechanism causally via natural experiment: within model$\times$task units, stochastic drift from the canonical path determines whether a capable agent succeeds or fails. The natural experiment is well-identified but observational in one important respect---it cannot rule out that some unobserved run-level variable simultaneously causes both higher adherence and higher success. A direct experimental test would close this gap by \emph{manipulating} canonical path adherence and measuring the downstream effect on success. The cleanest design would exploit Toolathlon's containerized, deterministic environment to branch trajectories at the drift detection point (75\% completion): one branch continues unmodified (control), while the other receives either a hard reset to a canonical tool prefix (strong intervention) or a mid-trajectory prompt nudge toward canonical tools (soft intervention). Comparing success rates across branches within the same model$\times$task unit would provide the experimental analog of our observational $+0.060$ Jaccard estimate, and directly validate the $+8.8$pp counterfactual lift reported in Section~\ref{sec:discussion}. We leave this experimental validation, and a systematic comparison of intervention designs, to follow-up work.

\paragraph{Generalization.}
Our causal identification is restricted to the 515 mixed-outcome units (22.5\% of all model$\times$task pairs) where outcomes vary across runs. However, canonical path adherence shows a monotonic gradient across all unit types: always-fail trajectories show mean adherence of 0.525, mixed-outcome failure runs 0.585, mixed-outcome success runs 0.644, and always-succeed trajectories 0.665 (all adjacent-group differences $p < 0.0001$). This four-level gradient suggests that the canonical path mechanism generalizes beyond marginal cases. The causal estimate of $+5.3$pp per $+0.060$ Jaccard is identified from mixed units, but the descriptive pattern and variance signatures (Section~\ref{sec:results}) are consistent across the full dataset. The breadth of the Toolathlon benchmark---604 tools spanning 32 applications from everyday productivity platforms to professional engineering environments---provides additional grounds against a ``toy task'' interpretation: canonical paths are detectable across a wide range of real-world task structures, not a narrow or artificially constrained tool ecosystem.

\section{Conclusion}
\label{sec:conclusion}

Our results suggest that agent failure is partially a reliability problem, not merely a capability problem. Tasks tend to have empirical consensus tool-use strategies that characterize successful behavior across models, and stochastic LLM sampling can cause capable agents to drift away from these strategies, producing failures despite sufficient underlying capability. Under the identifying assumption that run-level stochasticity is exogenous, the drift is gradual and cumulative rather than caused by single early mistakes, and the pattern holds across all 11 frontier model families we study.

The variance analysis (Section~\ref{sec:results}) further clarifies the boundary between failure types: always-succeed units show low adherence variance (the model reliably finds the canonical path), always-fail units show intermediate variance (the model consistently takes the wrong path, a capability signature), and mixed units show the highest variance (the model is capable but stochastically unreliable). Reliability interventions targeting canonical path adherence are therefore most valuable for mixed units---the population where failure is caused by drift, not by capability gaps that no amount of path guidance would fix.

These findings suggest a reframing of the agent improvement agenda. Alongside capability improvements (better models), the field may benefit from reliability improvements: extracting consensus tool strategies from successful runs, monitoring adherence during execution, and intervening when trajectories drift toward failure. The natural experiment embedded in repeated-run benchmarks provides a powerful and underutilized tool for identifying such mechanisms.

\bibliography{main}
\bibliographystyle{ACM-Reference-Format}

\appendix
\section{Sample Construction Flowchart}
\label{app:flowchart}

Figure~\ref{fig:flowchart} documents the full sample construction from all trajectories to the 488 units used in the headline CF-LOO analysis.

\begin{center}
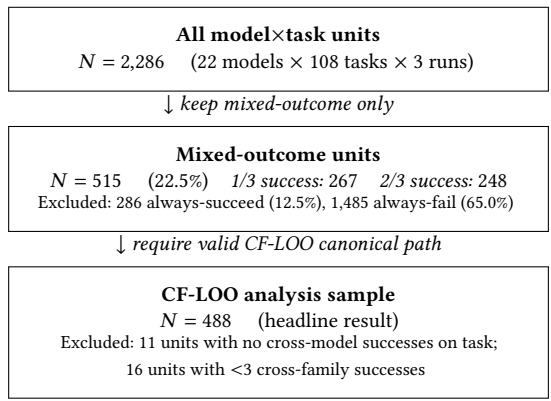

\small
\setlength{\tabcolsep}{0pt}
\begin{tabular}{c}
\fbox{\begin{minipage}{0.82\columnwidth}
\centering\vspace{4pt}
\textbf{All model$\times$task units}\\
$N = 2{,}286$ \quad (22 models $\times$ 108 tasks $\times$ 3 runs)
\vspace{4pt}
\end{minipage}}\\[3pt]
$\downarrow$ \textit{keep mixed-outcome only}\\[3pt]
\fbox{\begin{minipage}{0.82\columnwidth}
\centering\vspace{4pt}
\textbf{Mixed-outcome units}\\
$N = 515$ \quad (22.5\%)\quad
\textit{1/3 success:} 267\quad\textit{2/3 success:} 248\\[-2pt]
{\footnotesize Excluded: 286 always-succeed (12.5\%),
1{,}485 always-fail (65.0\%)}
\vspace{4pt}
\end{minipage}}\\[3pt]
$\downarrow$ \textit{require valid CF-LOO canonical path}\\[3pt]
\fbox{\begin{minipage}{0.82\columnwidth}
\centering\vspace{4pt}
\textbf{CF-LOO analysis sample}\\
$N = 488$ \quad (headline result)\\[-2pt]
{\footnotesize Excluded: 11 units with no cross-model successes on
task;}\\ {\footnotesize 16 units with $<$3 cross-family successes}
\vspace{4pt}
\end{minipage}}
\end{tabular}
\captionof{figure}{Sample construction from full dataset to CF-LOO analysis
sample. The less conservative LOO specification retains $n=495$
units; results are substantively identical.}
\label{fig:flowchart}
\end{center}

\section{Translating Adherence Gaps to Success Probability}
\label{app:success_lift}

Table~\ref{tab:main} reports within-unit adherence gaps in Jaccard similarity units, which are methodologically precise but not immediately intuitive. Table~\ref{tab:success_prob} translates each specification's adherence gap into estimated success probability lift using a logistic regression model fit on all mixed-outcome trajectories ($n = 1{,}476$).

The model regresses success on within-unit demeaned canonical path adherence, controlling for trajectory-level heterogeneity. The marginal effect (evaluated at mean adherence) converts a Jaccard gap into percentage-point change in success probability. For the headline CF-LOO specification, the $+0.060$ Jaccard gap translates to a $+5.3$pp success probability lift (odds ratio $= 1.24$, Cohen's $d = 0.48$, a medium effect).

Two findings emerge. First, the success probability lift is stable across robustness specifications, ranging from $+3.2$pp (generic tools only) to $+7.2$pp (naive), mirroring the stability of the Jaccard gaps themselves. This confirms that the practical impact is robust to alternative canonical path definitions and controls.

Second, even the most conservative specification (generic tools only, which excludes all domain-specific tools and yields the smallest Jaccard gap) still implies a meaningful $+3.2$pp lift---comparable to the difference between adjacent model versions in capability benchmarks. This underscores that canonical path adherence has practical predictive power beyond statistical significance.

\begin{center}
\small
\begin{tabular}{lccc}
\toprule
\textbf{Specification} & \textbf{Jaccard Gap} & \textbf{Success Lift}
& \textbf{OR} \\
\midrule
Original (naive)         & +0.082 & +7.2pp & 1.34 \\
Leave-one-out            & +0.067 & +5.9pp & 1.27 \\
CF-LOO (headline)        & +0.060 & +5.3pp & 1.24 \\
Generic tools only       & +0.036 & +3.2pp & 1.14 \\
Length-residualized      & +0.053 & +4.7pp & 1.21 \\
Leave-one-task-out       & +0.050 & +4.4pp & 1.19 \\
\bottomrule
\end{tabular}
\captionof{table}{Success probability lift estimates for each robustness specification. The lift is the estimated percentage-point increase in success probability associated with the within-unit adherence gap from Table~\ref{tab:main}, computed via logistic regression on within-unit demeaned adherence (coefficient $= 3.549$, fit on CF-LOO sample). OR = odds ratio.}
\label{tab:success_prob}
\end{center}


\section{Full Robustness Specification Table}
\label{app:robustness}

Table~\ref{tab:full_robust} reports all robustness specifications with full details. The first three rows correspond to progressively stricter canonical path definitions (naive, LOO, CF-LOO). Rows R3--R5 test alternative explanations. Row R6 reports threshold sensitivity across four canonical path definitions. Rows E1--E3 report the three exogeneity placebo tests supporting the identifying assumption; E2 is split into three rows, one per trajectory checkpoint tested.

\begin{table*}[t]
\centering
\caption{Full robustness and placebo specification table. For robustness specifications (--/R/R6 rows), the Statistic column reports the within-unit success$-$failure Jaccard adherence gap. For E1, it reports the minority first-tool-choice success rate (binomial test vs.\ $H_0{=}0.50$). For E2, it reports the within-unit adherence gap at that trajectory checkpoint. 95\% CIs are bootstrap percentile intervals (5,000 draws, clustered by unit) for Jaccard gaps; exact binomial CI for E1. $^{***}p<0.001$, $^{**}p<0.01$, $^{*}p<0.05$, ns\,$p\geq0.05$. E2 uses $n=454$ units (vs.\ $n=488$ headline) because the partial-trajectory analysis excludes trajectories with fewer than 4 tool calls and those lacking sufficient cross-family successful runs at each checkpoint. E1 uses $n=308$ minority-tool observations from 222 units where first tool varies across runs.}
\label{tab:full_robust}
\small
\setlength{\tabcolsep}{5pt}
\begin{tabularx}{\linewidth}{llcccccX}
\toprule
\textbf{ID} & \textbf{Specification} & \textbf{Estimate} &
\textbf{95\% CI} & \textbf{$N$} & \textbf{Sig.} &
\textbf{What it tests} \\
\midrule
\multicolumn{7}{l}{\textit{Canonical path definition}} \\
-- & Naive (all models) & +0.082 & [+0.059,~+0.105] & 505 &
  $^{***}$ & Baseline; circular by construction \\
R1 & Leave-one-out (LOO) & +0.067 & [+0.050,~+0.084] & 495 &
  $^{***}$ & Removes target model from path definition \\
R2 & CF-LOO (headline) & +0.060 & [+0.043,~+0.077] & 488 &
  $^{***}$ & Removes entire model family \\
\midrule
\multicolumn{7}{l}{\textit{Alternative explanations}} \\
R3 & Generic tools only & +0.036 & [+0.022,~+0.051] & 473 &
  $^{***}$ & Excludes domain-obvious tools \\
R4 & Length-residualized & +0.053 & [+0.037,~+0.071] & 481 &
  $^{***}$ & Residualizes adherence on trajectory length \\
R5 & Leave-one-task-out & +0.050 & [+0.033,~+0.066] & 471 &
  $^{***}$ & Removes most influential task (\texttt{set-conf-cr-ddl}) \\
\midrule
\multicolumn{7}{l}{\textit{R6: Canonical threshold sensitivity}} \\
R6a & Threshold 40\% & +0.057 & [+0.041,~+0.075] & 488 &
  $^{***}$ & More inclusive canonical path \\
R6b & Threshold 50\% & +0.060 & [+0.043,~+0.077] & 488 &
  $^{***}$ & Headline specification \\
R6c & Threshold 60\% & +0.062 & [+0.045,~+0.079] & 488 &
  $^{***}$ & More restrictive canonical path \\
R6d & Threshold 70\% & +0.055 & [+0.039,~+0.073] & 488 &
  $^{***}$ & Most restrictive canonical path \\
\midrule
\multicolumn{7}{l}{\textit{Exogeneity placebo tests (E1--E3)}} \\
E1 & First tool call distribution & $51.3\%$ & [45.6\%,~57.0\%] & 308 &
  ns ($p=0.69$) & Minority first-tool success rate $\approx 50\%$; binomial vs.\ $H_0{=}0.50$ \\
E2 & Gradual drift null at 10\% & $-0.002$ & [$-0.013$,~$+0.009$] & 454 &
  ns ($p=0.74$) & Gap indistinguishable from 0 at 10\% of trajectory \\
E2 & Gradual drift null at 25\% & $+0.011$ & [$-0.003$,~$+0.024$] & 454 &
  ns ($p=0.12$) & Gap indistinguishable from 0 at 25\% of trajectory \\
E2 & Gradual drift null at 50\% & $+0.012$ & [$-0.002$,~$+0.026$] & 454 &
  ns ($p=0.09$) & Gap indistinguishable from 0 at 50\% of trajectory \\
E3 & CF-LOO independence & \multicolumn{2}{c}{n/a --- structural} & 488 &
  structural & Within-family correlation excluded by construction \\
\bottomrule
\end{tabularx}
\end{table*}

\section{Per-Model Reliability Metrics}
\label{app:reliability_metrics}

Table~\ref{tab:reliability} applies the mixed-outcome fraction metric to all 22 Toolathlon models. For each model we report four metrics: P@1 (mean success rate per trajectory), P@3 (fraction of tasks with at least one success across 3 runs---the capability ceiling), P\^{}3 (fraction of tasks succeeding on all 3 runs---the reliability floor), and MO/P@3 (mixed-outcome units as a fraction of P@3-capable tasks).

MO/P@3 is the \emph{intervention scope} metric: of the tasks a model is capable of solving, what fraction does it solve unreliably? This is the upper bound on the performance gain that reliability interventions---such as canonical path scaffolding---can recover, since always-fail units represent capability gaps that path guidance cannot address.

Two findings emerge from Table~\ref{tab:reliability}. First, MO/P@3 is inversely correlated with capability: weaker models not only solve fewer tasks but solve almost none of their capable tasks consistently (gemini-2.5-flash: 100\%; minimax-m2: 91.3\%; o4-mini: 85.7\%), while stronger models show substantially lower intervention scope (claude-4.5-opus: 46.8\%; gpt-5.1: 48.9\%). This suggests that reliability failures are disproportionately concentrated in weaker models---a pattern invisible in P@1 rankings.

Second, MO/P@3 separates models that appear equivalent on standard
metrics. Grok-4-fast and o3 are nearly tied at P@1 $\approx$ 19\%, but their MO/P@3 scores diverge sharply: 82.9\% vs.\ 61.5\%. If canonical path scaffolding recovers a fixed fraction of mixed-outcome units, grok-4-fast has proportionally more to gain from the intervention---making MO/P@3 a useful planning metric when choosing between similarly-capable models for reliability-targeted deployment.

\begin{center}
\begin{minipage}{\columnwidth}
\centering
\small
\setlength{\tabcolsep}{5pt}
\begin{tabular}{lrrrr}
\toprule
\textbf{Model} & \textbf{P@1} & \textbf{P@3} & \textbf{P\^{}3} &
\textbf{MO/P@3} \\
\midrule
\multicolumn{5}{l}{\textit{Proprietary models}} \\
\texttt{claude-4.5-opus}       & 43.5 & 57.4 & 30.6 & 46.8\% \\
\texttt{claude-4.5-sonnet}     & 38.6 & 51.9 & 20.4 & 60.7\% \\
\texttt{gemini-3-pro-preview}  & 36.4 & 48.1 & 23.1 & 51.9\% \\
\texttt{gpt-5.1}               & 33.3 & 43.5 & 22.2 & 48.9\% \\
\texttt{gpt-5}                 & 32.7 & 46.0 & 18.0 & 60.9\% \\
\texttt{gpt-5-high}            & 32.6 & 45.7 & 19.6 & 57.1\% \\
\texttt{claude-4-sonnet}       & 29.9 & 41.7 & 17.6 & 57.8\% \\
\texttt{grok-4}                & 28.2 & 39.4 & 17.3 & 56.1\% \\
\texttt{claude-4.5-haiku}      & 26.2 & 39.8 & 13.0 & 67.4\% \\
\texttt{grok-code-fast-1}      & 19.2 & 31.7 &  9.6 & 69.7\% \\
\texttt{grok-4-fast}           & 18.9 & 33.0 &  5.7 & 82.9\% \\
\texttt{o3}                    & 18.6 & 26.8 & 10.3 & 61.5\% \\
\texttt{gpt-5-mini}            & 15.8 & 25.3 &  6.3 & 75.0\% \\
\texttt{o4-mini}               & 15.1 & 26.9 &  3.8 & 85.7\% \\
\texttt{gemini-2.5-pro}        & 10.5 & 21.3 &  2.8 & 87.0\% \\
\texttt{minimax-m2}            & 10.2 & 21.3 &  1.9 & 91.3\% \\
\texttt{gemini-2.5-flash}      &  3.5 &  7.6 &  0.0 & 100.0\% \\
\midrule
\multicolumn{5}{l}{\textit{Open-source models}} \\
\texttt{deepseek-3.2-thinking} & 35.2 & 54.6 & 16.7 & 69.5\% \\
\texttt{deepseek-v3.2-exp}     & 20.4 & 28.3 & 12.3 & 56.7\% \\
\texttt{glm-4.6}               & 18.8 & 29.6 &  9.3 & 68.8\% \\
\texttt{qwen-3-coder}          & 15.4 & 22.5 &  6.9 & 69.6\% \\
\texttt{kimi-k2-0905}          & 14.7 & 25.3 &  6.6 & 73.9\% \\
\bottomrule
\end{tabular}
\captionof{table}{Per-model reliability metrics for all 22 Toolathlon models,
sorted by P@1.
P@3 = capability ceiling (\% of tasks with $\geq$1 success);
P\^{}3 = reliability floor (\% of tasks succeeding all 3 runs);
MO/P@3 = mixed-outcome fraction among P@3-capable tasks, i.e.\ the
fraction of solvable tasks solved unreliably.
MO/P@3 is the upper bound on what reliability interventions can
recover per model.}
\label{tab:reliability}
\end{minipage}
\end{center}

\section{Per-Task Statistics}
\label{app:tasks}
\begin{table*}[t]
\caption{Per-task statistics for the 67 tasks with mixed-outcome units and valid CF-LOO estimates (3 of the 70 mixed-outcome tasks lack sufficient cross-family successful runs to construct a CF-LOO canonical path and are excluded here). Tasks sorted by canonical path strength (descending). Str.\ = canonical strength (mean pairwise Jaccard among successful runs); Eff.\ = within-unit adherence effect (success$-$failure); $N$ = mixed-outcome units; Succ.\ = overall success rate across all models. 54/67 tasks (80.6\%) show positive effects.}
\label{tab:tasks}
\small
\setlength{\tabcolsep}{4pt}
\begin{minipage}[t]{0.48\textwidth}
\centering
\begin{tabular}{lcccc}
\toprule
\textbf{Task} & \textbf{Str.} & \textbf{Eff.} & \textbf{$N$} & \textbf{Succ.} \\
\midrule
\texttt{interview-report} & 0.838 & +0.067 & 1 & 0.11 \\
\texttt{notion-find-job} & 0.798 & +0.067 & 5 & 0.15 \\
\texttt{landing-task-reminder} & 0.771 & +0.037 & 5 & 0.22 \\
\texttt{imagenet} & 0.763 & -0.009 & 6 & 0.21 \\
\texttt{sla-timeout-monitor} & 0.752 & +0.010 & 7 & 0.24 \\
\texttt{notion-hr} & 0.739 & +0.050 & 7 & 0.29 \\
\texttt{nhl-b2b-analysis} & 0.736 & +0.104 & 10 & 0.72 \\
\texttt{ppt-analysis} & 0.735 & +0.026 & 7 & 0.30 \\
\texttt{notion-movies} & 0.734 & +0.036 & 10 & 0.35 \\
\texttt{student-interview} & 0.729 & +0.129 & 7 & 0.12 \\
\texttt{canvas-art-manager} & 0.710 & +0.000 & 5 & 0.26 \\
\texttt{woocommerce-stock-alert} & 0.702 & +0.081 & 8 & 0.26 \\
\texttt{set-conf-cr-ddl} & 0.701 & +0.302 & 10 & 0.48 \\
\texttt{woocommerce-product-recall} & 0.700 & +0.174 & 7 & 0.45 \\
\texttt{payable-invoice-checker} & 0.694 & +0.055 & 9 & 0.57 \\
\texttt{git-repo} & 0.690 & -0.023 & 7 & 0.82 \\
\texttt{nvidia-stock-analysis} & 0.688 & +0.031 & 8 & 0.48 \\
\texttt{train-ticket-plan} & 0.677 & -0.009 & 7 & 0.76 \\
\texttt{excel-market-research} & 0.669 & +0.176 & 6 & 0.18 \\
\texttt{quantitative-financial-ana\ldots} & 0.667 & +0.175 & 6 & 0.14 \\
\texttt{canvas-new-students-notifi\ldots} & 0.666 & +0.005 & 4 & 0.20 \\
\texttt{k8s-safety-audit} & 0.665 & +0.067 & 6 & 0.27 \\
\texttt{canvas-do-quiz} & 0.664 & -0.017 & 7 & 0.45 \\
\texttt{price-comparison} & 0.663 & -0.021 & 3 & 0.85 \\
\texttt{personal-website-construct} & 0.655 & +0.066 & 9 & 0.36 \\
\texttt{academic-pdf-report} & 0.652 & +0.107 & 7 & 0.54 \\
\texttt{game-statistics} & 0.651 & +0.017 & 12 & 0.52 \\
\texttt{git-bug-hunt} & 0.651 & +0.197 & 6 & 0.80 \\
\texttt{machine-operating} & 0.651 & +0.092 & 4 & 0.37 \\
\texttt{fillout-online-forms} & 0.647 & +0.009 & 7 & 0.61 \\
\texttt{apply-phd-email} & 0.645 & +0.069 & 10 & 0.38 \\
\texttt{invoice-org} & 0.637 & +0.076 & 6 & 0.32 \\
\texttt{dietary-health} & 0.633 & +0.089 & 8 & 0.15 \\
\texttt{canvas-art-quiz} & 0.629 & +0.090 & 6 & 0.73 \\
\bottomrule
\end{tabular}
\end{minipage}%
\hfill
\begin{minipage}[t]{0.48\textwidth}
\centering
\begin{tabular}{lcccc}
\toprule
\textbf{Task} & \textbf{Str.} & \textbf{Eff.} & \textbf{$N$} & \textbf{Succ.} \\
\midrule
\texttt{canvas-homework-grader-pyt\ldots} & 0.621 & +0.037 & 10 & 0.36 \\
\texttt{trip-adviser} & 0.620 & -0.020 & 7 & 0.35 \\
\texttt{woocommerce-new-welcome} & 0.618 & +0.023 & 6 & 0.12 \\
\texttt{upenn-campus-route} & 0.613 & +0.093 & 14 & 0.40 \\
\texttt{courses-ta-hws} & 0.601 & +0.089 & 9 & 0.20 \\
\texttt{gdp-cr5-analysis} & 0.597 & +0.086 & 2 & 0.09 \\
\texttt{course-schedule} & 0.596 & +0.042 & 4 & 0.08 \\
\texttt{k8s-redis-helm-upgrade} & 0.593 & +0.098 & 2 & 0.12 \\
\texttt{verl-dataset} & 0.591 & +0.022 & 13 & 0.45 \\
\texttt{git-milestone} & 0.586 & +0.181 & 3 & 0.89 \\
\texttt{dataset-license-issue} & 0.578 & +0.005 & 10 & 0.30 \\
\texttt{stock-build-position} & 0.571 & +0.212 & 10 & 0.33 \\
\texttt{excel-data-transformation} & 0.565 & -0.044 & 13 & 0.45 \\
\texttt{sales-accounting} & 0.565 & +0.171 & 9 & 0.32 \\
\texttt{woocommerce-update-cover} & 0.560 & -0.009 & 4 & 0.36 \\
\texttt{academic-warning} & 0.559 & +0.166 & 7 & 0.25 \\
\texttt{shopping-helper} & 0.555 & +0.000 & 9 & 0.17 \\
\texttt{cooking-guidance} & 0.549 & -0.019 & 10 & 0.28 \\
\texttt{reimbursement-form-filler} & 0.549 & -0.042 & 6 & 0.11 \\
\texttt{wandb-best-score} & 0.547 & +0.007 & 5 & 0.33 \\
\texttt{find-alita-paper} & 0.529 & +0.313 & 9 & 0.73 \\
\texttt{k8s-deployment-cleanup} & 0.525 & +0.364 & 1 & 0.17 \\
\texttt{inventory-sync} & 0.523 & -0.017 & 5 & 0.25 \\
\texttt{flagged-transactions} & 0.513 & -0.020 & 6 & 0.33 \\
\texttt{wandb-shortest-length} & 0.506 & +0.097 & 11 & 0.38 \\
\texttt{trip-itinerary-generator} & 0.505 & +0.065 & 9 & 0.36 \\
\texttt{canvas-submit-late-work} & 0.475 & +0.061 & 14 & 0.64 \\
\texttt{huggingface-upload} & 0.449 & +0.054 & 6 & 0.18 \\
\texttt{logical-datasets-collection} & 0.432 & +0.048 & 9 & 0.32 \\
\texttt{language-school} & 0.418 & +0.059 & 5 & 0.12 \\
\texttt{add-bibtex} & 0.398 & +0.005 & 4 & 0.14 \\
\texttt{subway-planning} & 0.393 & -0.115 & 12 & 0.52 \\
\texttt{ipad-edu-price} & 0.377 & +0.223 & 13 & 0.68 \\
\midrule
\textit{Mean / Total} & \textit{0.616} & \textit{+0.068} & \textit{488} & \textit{0.38} \\
\bottomrule
\end{tabular}
\end{minipage}
\end{table*}

Table~\ref{tab:tasks} reports canonical path strength, within-unit adherence effect, number of mixed-outcome units, and overall task success rate for all 67 tasks with at least one mixed-outcome unit. Canonical strength is the mean pairwise Jaccard similarity among successful runs on the task across all models. Effect size is the mean within-unit success$-$failure adherence gap for that task's mixed-outcome units.

\section{Illustrative Trajectory Example}
\label{app:example}

Figure~\ref{fig:example} shows the tool-call sequences for one mixed-outcome unit (\texttt{git-milestone}, model \texttt{glm-4.6}). The task requires creating a GitHub milestone from a structured data source. The CF-LOO canonical path consists of three tools:
\texttt{fetch-fetch\_json},
\texttt{filesystem-write\_file}, and
\texttt{local-claim\_done}.

The successful run (Run~2, left) completes in 6 steps. The representative failure run (Run~1, right) makes identical choices for the first 5 steps, then switches to \texttt{fetch\_html} at step~6 --- the divergence point --- and never recovers, spiraling through 32 further off-canonical calls before terminating without writing the file. Each deviation shifts the agent's context state in ways that make the canonical \texttt{write\_file} conclusion progressively less likely, illustrating the compounding drift mechanism. The within-unit adherence gap for this unit ($+0.694$ Jaccard) is well above the sample mean ($+0.060$).

\clearpage
\onecolumn
\begin{center}
\begin{tikzpicture}[
  node distance=0.18cm,
  header/.style={font=\small\bfseries},
  subheader/.style={font=\scriptsize\itshape, text=gray!70!black},
  divlabel/.style={font=\scriptsize\bfseries, text=BrickRed},
  toolbox/.style={
    rounded corners=3pt, minimum width=3.0cm, minimum height=0.42cm,
    inner sep=4pt, font=\ttfamily\scriptsize, text width=2.9cm,
    draw, align=left
  },
  canonical/.style={toolbox, draw=teal, fill=teal!10, text=teal!80!black},
  offcanon/.style={toolbox, draw=BrickRed, fill=BrickRed!10,
    text=BrickRed!80!black},
  outcomebox/.style={rounded corners=4pt, minimum width=3.0cm,
    minimum height=0.5cm, font=\small\bfseries, align=center, inner sep=4pt},
]

\node[header] (lhead) at (0,0)     {Run 2 — Success};
\node[subheader, below=0.04cm of lhead] (lsub) {adh\,=\,1.00 $|$ 6 steps};

\node[header] (rhead) at (4.6,0)   {Run 1 — Failure};
\node[subheader, below=0.04cm of rhead] (rsub) {adh\,=\,0.11 $|$ 38 steps};

\node[canonical, below=0.28cm of lsub] (l1) {1.\ \ fetch-fetch\_json};
\node[canonical, below=0.18cm of l1]   (l2) {2.\ \ fetch-fetch\_json};
\node[canonical, below=0.18cm of l2]   (l3) {3.\ \ fetch-fetch\_json};
\node[canonical, below=0.18cm of l3]   (l4) {4.\ \ fetch-fetch\_json};

\node[canonical] at (l1 -| rhead)    (r1) {1.\ \ fetch-fetch\_json};
\node[canonical, below=0.18cm of r1] (r2) {2.\ \ fetch-fetch\_json};
\node[canonical, below=0.18cm of r2] (r3) {3.\ \ fetch-fetch\_json};
\node[canonical, below=0.18cm of r3] (r4) {4.\ \ fetch-fetch\_json};

\node[canonical, below=0.18cm of l4] (l5) {5.\ \ filesystem-write\_file};
\node[canonical, below=0.18cm of r4] (r5) {5.\ \ fetch-fetch\_json};

\node[canonical, below=0.18cm of l5] (l6) {6.\ \ local-claim\_done};
\node[offcanon,  below=0.18cm of r5] (r6) {6.\ \ fetch-fetch\_html};

\node[divlabel, right=0.12cm of r6] (divlbl) {$\leftarrow$ divergence};

\node[outcomebox, below=0.28cm of l6,
      draw=teal, fill=teal!15, text=teal!80!black]
      (lout) {\checkmark\ PASS};

\node[offcanon, below=0.18cm of r6]  (r7)  {7.\ \ fetch-fetch\_html};
\node[offcanon, below=0.18cm of r7]  (r8)  {8--12.\ search\_in\_turn ×5};
\node[offcanon, below=0.18cm of r8]  (r9)  {13--21. search\_overlong ×9};
\node[offcanon, below=0.18cm of r9]  (r10) {22--28. local-web\_search ×7};
\node[offcanon, below=0.18cm of r10] (r11) {29--32. fetch-fetch\_html ×4};
\node[offcanon, below=0.18cm of r11] (r12) {33--38. local-web\_search ×6};

\node[outcomebox, below=0.28cm of r12,
      draw=BrickRed, fill=BrickRed!12, text=BrickRed!80!black]
      (rout) {$\times$\ FAIL — no write\_file};

\begin{scope}[on background layer]
  \node[fill=sharedgray, rounded corners=4pt,
        fit=(l1)(l4)(r1)(r4), inner sep=4pt,
        label={[font=\scriptsize\itshape,
                text=gray!60!black]left:shared prefix}] {};
\end{scope}

\draw[gray!40, thick] (2.25, 0.1) -- (2.25, -9.6);

\node[below=0.5cm of rout, xshift=-2.3cm,
      font=\scriptsize, align=left] (leg) {%
  \textcolor{teal}{$\blacksquare$}~canonical tool\quad
  \textcolor{BrickRed}{$\blacksquare$}~off-canonical tool
};

\end{tikzpicture}

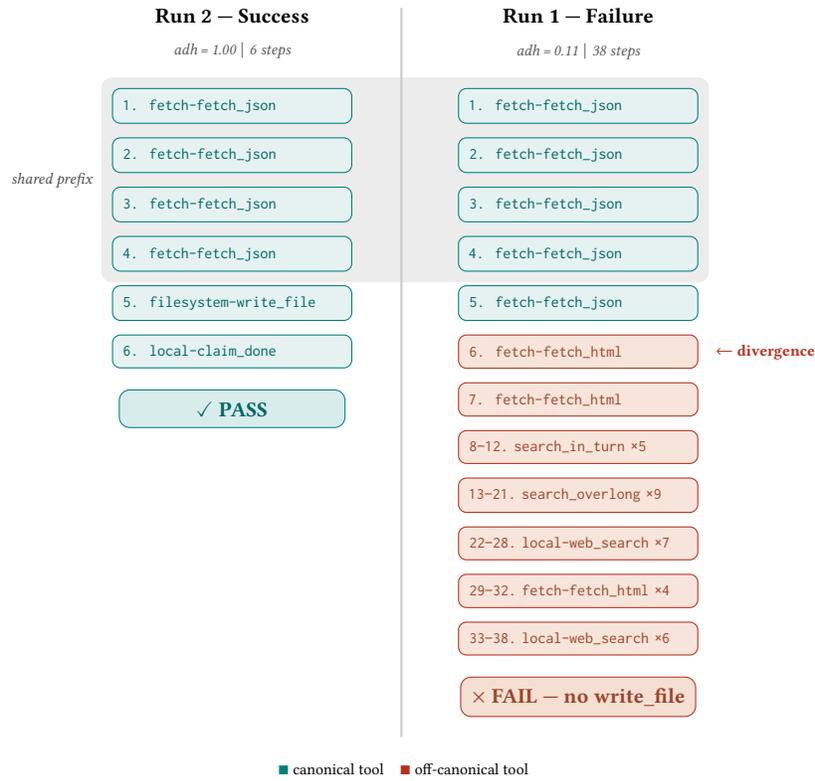
\captionof{figure}{Trajectory comparison for \texttt{git-milestone} (\texttt{glm-4.6}).
  Both runs share an identical four-step prefix (shaded); Run~2 then
  completes canonically in 2 further steps, while Run~1 deviates at
  step~6 and spirals through 32 off-canonical calls without reaching
  \texttt{write\_file}. Within-unit adherence gap: $+0.694$ Jaccard
  (sample mean: $+0.060$).}
\label{fig:example}
\end{center}
\twocolumn

\end{document}